\documentclass[10pt,twocolumn,journal]{IEEEtran}
\usepackage{graphicx,amssymb,amsmath,amsthm}
\usepackage{verbatim}
\newcommand{\VS}{{\cal{S}}} 

\newcommand{\la}{\langle}
\newcommand{\ra}{\rangle}

\newcommand{\Spann}{{\mbox{\rm{span}}}}
\newcommand{\til}{\tilde}

\newtheorem{remark}{Remark}

\def\be{\begin{equation}}
\def\ee{\end{equation}}
\def\ben{\begin{eqnarray}}
\def\een{\end{eqnarray}}
\def\key{\mathbf{key}}
\def\seed{\mathbf{seed}}
\def\R{\mathbb{R}}

\def\vI{\mathbf{I}}
\def\vv{\mathbf{v}}
\def\ve{\mathbf{e}}
\def\vd{\mathbf{d}}
\def\vo{\mathbf{o}}
\def\vot{\til{\mathbf{o}}}
\def\vu{\mathbf{u}}
\def\vy{\mathbf{y}}

\def\vR{\mathbf{R}}
\def\vF{\mathbf{F}}
\def\vG{\mathbf{G}}

\title{Sparsity and `Something Else': An Approach to Encrypted Image 
Folding}
\author{James~Bowley and Laura~Rebollo-Neira\\
Mathematics Department, Aston University, Birmingham B4 7ET, UK}

\begin{document}
\maketitle
\begin{abstract}
A property of sparse representations in relation to their capacity 
for information storage is discussed.
It is shown that this feature can be used 
for an application that we term Encrypted Image Folding.
The proposed procedure is realizable through any suitable
transformation. In particular, in this paper we illustrate the 
approach by recourse to the Discrete Cosine Transform and a 
combination of redundant Cosine and Dirac dictionaries. 
The main advantage of the proposed technique is that both storage 
and encryption can be achieved simultaneously using simple processing steps.
\end{abstract}
\section{Introduction} 
The problem of reducing the dimensionality of a piece of 
data without losing the information content is of paramount importance 
in signal processing.
Well-established transforms, from classical Fourier and Cosine
Transforms to Wavelets,  Wavelet Packets, and  Lapped 
Transforms, just to mention the most popular ones, are usually 
applied for generating the transformed domain where the processing 
tasks are realized.
Signals amenable to transformation into 
data sets of smaller cardinality are said to be compressible. 
Natural images, for instance, provide a typical example 
of compressible data. 

In the last fifteen years emerging techniques for signal 
representation are addressing the matter by means of highly nonlinear 
methodologies which decompose the signal into a superposition 
of vectors, normally called `atoms',  selected from a large 
redundant set called a `dictionary'. The representation 
qualifies to be sparse if the number of atoms for a satisfactory 
signal approximation is considerably smaller than the dimension 
of the original data. 
Available methodologies for highly nonlinear approximations are known 
as Pursuit Strategies. This comprises Basis Pursuit \cite{CDS98,DT05} 
and Matching-Pursuit-like algorithms, including
Orthogonal Matching Pursuit (OMP) and variations of these methods 
\cite{MZ93,PRK93,RL02,AR06,JVF06,DTD06,NV07,NT09}. 
The other ingredient of highly nonlinear approximations is, 
of course, the dictionary providing the atoms for the selection.
In this respect, Gabor dictionaries have been shown to be
useful for image and video processing \cite{FCR06,FVF06}. Combined 
 dictionaries, arising by merging for instance
orthogonal bases, have received consideration in relation to 
the theoretical analysis of Pursuit Strategies 
\cite{DH01,EB02,FN03,GN03,Tro04}. From a different perspective, 
other approaches are based on dictionaries learned from large 
data sets \cite{OF97,AEB06}.

This communication exploits an inherent side-effect of 
sparse representations. Since sparsity entails 
a projection onto a subspace of lower dimensionality, a 
null space is generated. Extra information can be embedded in such a 
space and then stably extracted.
In particular, we discuss an application involving the null space 
yielded by the sparse representation of an image, to store part of
the image itself in encrypted form.
We term this application Encrypted Image Folding (EIF). The main advantage 
of this proposal, in relation to standard techniques, is that 
{\em {storage and encryption can be achieved simultaneously
by means of simple data processing steps}}. The proposed procedure 
can be carried out through any suitable transformation. In 
particular, we consider here
  the Discrete Cosine Transform (DCT) and 
a mixed dictionary composed of a Redundant 
Discrete Cosine (RDC) dictionary and a discrete Dirac Basis (DB).
RDC and DB dictionaries are considered separately in \cite{CDS98}. 
A theoretical discussion with regards to a random collection of 
elements of a Discrete Sine basis and a DB is presented in \cite{Tro08}.
In this letter we would simply like to draw attention to the suitability of 
mixed dictionaries composed of RDC and DB, for image representation.
As far as sparsity is concerned, at the visually acceptable 
level of 40dB PSNR, they may render a significant improvement
in comparison to established fast transforms such as the
DCT and Wavelet Transform (WT).
An additional advantage of these dictionaries is that Matching
Pursuit-like strategies for selecting the atoms can be implemented at 
a reduced complexity cost by means of the DCT. For these reasons, 
we illustrate our approach for EIF
using a mixed RDC-DB dictionary, in addition to standard the DCT. 

The paper is organized as follows: Sec.~\ref{dictionaries} motivates the 
use of a mixed RDC-DB dictionary within the present framework.
Sec.~\ref{se} discusses the fact that a sparse representation can 
be used for embedding information. Based on such a possibility, 
a scheme for image folding and a simple encryption 
procedure, fully implementable by data processing, 
are discussed in Sec.\ref{eif}.
The conclusions are presented in Sec.~\ref{conclu}. 

\section{Sparse image representation by RDC-DB Dictionaries}
\label{dictionaries}
Let us start by introducing  the dictionaries and methodology which 
will be used in Section~\ref{se} for illustrating the present approach. 
Consider the set $D_a$ defined as 
$$D_a=\{\vv_i;\,v_{j,i}= p_i\cos(\frac{\pi(2j-1)(i-1)}{2M}),\,j=1,\ldots,N 
\}_{i=1}^M,$$
with $p_i,\,i=1,\ldots,M$ normalization factors and the notation 
$v_{j,i}$ indicating the component $j$ of vector $\vv_i \in \R^N$.
If $M=N$ this set is 
a Discrete Cosine (DC) orthonormal basis for 
$\R^N$. If $M=2lN$, with $l$ a positive integer, 
the set is a DC dictionary with redundancy $2l$. 

We further consider the set $D_b$, which is a 
discrete DB, also known as standard orthonormal basis i.e., 
$$D_b=\{\ve_i \in \R^N ;\, e_{j,i}=\delta_{i,j},\,j=1,\ldots,N \}_{i=1}^N,$$
where $\delta_{i,j}=1$ if $i=j$ and zero otherwise.  
From the joint dictionary $D_{ab}= D_a \cup D_b$ a 
redundant dictionary $D$ for $\R^{N \times N}$ is obtained as 
the Kronecker product $D=D_{ab} \otimes D_{ab}$. We denote  by 
$\vd_n \in \R^{N \times N},\,n=1,\ldots,J$, where $J=(M+N)^2$,
the elements of dictionary 
$D$ and use them to construct
the atomic decomposition of an 
image $\vI \in \R^{N \times N}$ as 
\be
\label{atomic}
\vI^{K}= \sum_{i}^K c_i^K \vd_{\ell_i}.
\ee
The atoms $\vd_{\ell_i},\,i=1,\ldots,K$ are to be  
 selected from the dictionary $D$ by a Pursuit Strategy. 
In the examples we give here we have used OMP, which evolves as follows:
Setting $\vR^0=\vI$ at iteration $k+1$ the OMP algorithm
selects the atom, $\vd_{\ell_{k+1}}$ say, as the one maximazing the
absolute value of the Frobenius inner products
$\la \vd_i,\vR^{k} \ra_{F},\, i=1,\ldots,J$, i.e.,
\be
{\ell_{k+1}}=\arg\max\limits_{i=1,\ldots,J}|\la \vd_i ,\vR^{k} \ra_F|
\,\text{with}\, 
\vR^{k}= \vI - \sum_{i=1}^{k} c_i^k \vd_{\ell_i}. 
\label{omp}
\ee
The coefficients $c_i^k,\, i=1,\ldots,k$ in \eqref{omp}
are such that the  Frobenius norm $\|\vR^{k}\|_F$ is minimum. Our
implementation is based on Gram Schmidt orthonormalization and 
adaptive biorthogonalization, as proposed in \cite{RL02}.
The complexity is dominated by the calculation 
of the quantities $\la \vd_i , \vR^{k} \ra_F,\, i=1,\ldots,J$ in
\eqref{omp} at each iteration step. 
For the present dictionaries these quantities can be evaluated by 
fast DCT. In order to discuss the matter let us re-name the 
dictionary atoms as follows
\begin{align*}
\text{for } n & = 1,\ldots,M^2 \\
& \vd_n \to   \vv_i\otimes \vv_j, i= 1,\ldots,M,\, j=1,\ldots,M \\
\text{for } n & = M^2 +1,\ldots, M^2+MN \\
& \vd_n \to \vv_i\otimes \ve_j, i=1,\ldots,M,\, j=1,\ldots,N \\
\text{for } n & = M^2+ MN +1,\ldots, M^2+ 2MN \\
& \vd_n \to \ve_i\otimes \vv_j,  i=1,\ldots,N,\, j=1,\ldots,M \\
\text{for } n & = M^2 +2MN+1,\ldots, J \\
& \vd_n \to \ve_i\otimes \ve_j, i=1,\ldots,N,\, j=1,\ldots,N.
\end{align*}
Hence, by denoting as $R^k(s,r)$ the  element $(s,r)$ of matrix
$\vR^k$ and defining $\psi_{j,i}=\cos(\frac{\pi(2j-1)(i-1)}{2M})$, 
the inner products $\la \vd_i,\vR^{k} \ra_F,\, i=1,\ldots,J$
are calculated as
\begin{align}
\la  \vv_i\otimes \vv_j,\vR^k\ra_F & = p_i p_j\sum_{s,r=1}^N R^k(s,r)\psi_{s,i}\psi_{r,j} \label{eqq1}\\
\la  \vv_i\otimes \ve_j,\vR^k\ra_F & = p_i \sum_{s=1}^N R^k(s,j) \psi_{s,i}\label{eqq2}\\
\la  \ve_i\otimes \vv_j,\vR^k\ra_F & = p_j \sum_{r=1}^N R^k(i,r) \psi_{r,j}\label{eqq3}\\
\la  \ve_i\otimes \ve_j,\vR^k\ra_F & = R^k(i,j). \label{eqq4}
\end{align}
If $M=N$ \eqref{eqq1} is 
the 2D DCT of the residual $\vR^k$ whilst \eqref{eqq2}
and \eqref{eqq3} are the 1D DCT of the rows and columns 
of $\vR^k$, respectively.
If $M=2lN$, for some positive integer $l$, the calculations 
can also be carried out through fast DCT by 
zero padding. Thus, the 
complexity required for evaluation of inner products in 
\eqref{omp} 
is $O(M^2 \log_2 M)$.
\begin{table}
\begin{center}
\begin{tabular}{ | l || c | c | c | c | c | }
\hline
Image & Dictionary & DCT & DWT\\ \hline \hline
Barbara & 7.09 & 4.05    & 3.92   \\ \hline
Boat &  6.03 & 3.63 & 3.65 \\ \hline
Bridge & 3.70 & 2.06 & 2.20 \\ \hline
Film Clip &  8.06 & 4.53 & 4.81 \\ \hline
Jester &  6.28  &  3.6  &   3.88  \\ \hline
Lena &  10.06 &  6.50 & 6.97 \\ \hline
Mandrill & 3.32 & 1.91 & 1.90 \\ \hline
Peppers  & 7.74 & 4.36 & 3.39 \\ \hline
Photo (Fig~1) & 5.28 & 3.01  & 3.15   \\ \hline
\end{tabular}
\caption{Sparsity Ratio (for PSNR of 40dB)
achieved by the mixed RDC-DB dictionary 
and that yielded by DCT and DWT.} 
\label{table:cr}
\end{center}
\end{table}
In order to highlight the capacity of RDC-DB dictionaries to achieve 
sparse representation of natural images, we use them to represent  
the popular test images which are listed in the first column of 
Table \ref{table:cr} and the photo of Bertrand Russell shown in Fig~1.
For the actual processing we divide each image into 
blocks of $16 \times 16$ pixels. The sparsity measure we use 
is the Sparsity Ratio (SR) defined as
$${\text{SR}}=\frac{\text{total number of pixels}}{\text{total number of 
coefficients}}.$$
In all the cases the number of coefficients are determined 
so as to produce a PSNR of 40dB in the image reconstruction
and the dictionary is a mixed RDC (redundancy 2) and DB.
The results are given in the second column of Table \ref{table:cr}. 
For comparison the third column of this table shows  
results produced by DCT implemented using the same blocking 
scheme. For further comparison the results produced by 
the Cohen-Daubechies-Feauveau 9/7 DWT
(applied on the whole image at once) are displayed in the 
last column of Table \ref{table:cr}. Notice that, while for the fixed
 PSNR  of 40 dB the DCT and DWT yield comparable SR,
the corresponding SR obtained by the mixed dictionaries,
for all the images, is significantly higher. This motivates 
the use of RDC-DB dictionaries in the application we are 
proposing.
\section{Room for information embedding}
\label{se}
Since a sparse representation involves a projection onto a lower 
dimension subspace, it also creates room for storing
`something else'. The subspace, say $\VS_K$, 
spanned by the $K$-dictionary's atoms
$\{\vd_{\ell_i}\}_{i=1}^K$ rendering a sparse representation
of an image is a
 proper subspace of the image space $\R^{N \times N}$. Thus,
denoting by $\VS_K^\bot$ the orthogonal complement of $\VS_K$ in
$\R^{N \times N}$ we have $\R^{N \times N}= 
\VS_K \oplus^\bot \VS_K^\bot$ where 
$\oplus^\bot$ indicates orthogonal sum. Hence, if we  take an element  
$\vF\in \VS_K^\bot$ and add it to the image forming
 $\vG = \vI + \vF$, the  image $\vI $ can be recovered
from $\vG$ 
through the operation
\be
\label{coe}
{P}_{\VS_{K}} \vG= {P}_{\VS_{K}}( \vI + \vF)= \vI,
\ee
where ${P}_{\VS_{K}}$ is the orthogonal projection matrix, onto 
the subspace $\VS_{K}$.
This suggests the possibility of using the sparse representation 
of an image to embed the image with additional information 
stored in a matrix $\vF \in  \VS_K^\bot$. 
In order to do this, we apply the earlier proposed scheme 
to embed redundant 
representations \cite{MR04}, which in this case operates as 
described below.

{\bf{Embedding Scheme:}}
Consider that $\vI^K$ as in \eqref{atomic} is the reconstruction 
of a sparse representation of an image $\vI$.
We embed  $L=N^2-K$ numbers ${{h_i,\,i=1,\ldots,L}}$
into a matrix $\vF \in  \VS_K^\bot$ as prescribed below.
\begin{itemize}
\item
Take an orthonormal basis $\vu_i,\,i=1,\ldots,L$ for ${\VS_K^\bot}$ and
form matrix $\vF$ as the linear combination
\be
\label{plain}
\vF= \sum_{i=1}^L {h_i} \vu_i.
\ee
\item
Add $\vF$ to $\vI^K$ to obtain $\vG= \vI^K+ \vF.$
\end{itemize}

{\bf{Information Retrieval:}}
Given $\vG$ retrieve
the numbers ${{h_i,\,i=1,\ldots,L}}$ as follows.
\begin{itemize}
\item
Construct an orthogonal projection matrix ${P}_{\VS_{K}}$, onto
the subspace ${\VS_{K}}=\Spann\{\vd_{\ell_i}\}_{i=1}^{K}$ and
extract the image $\til{\vI}^K$  from $\vG$ as  
$\til{\vI}^K= {P}_{\VS_{K}} \vG$.
\item 
From the given $\vG$ and the extracted 
$\til{\vI}^K$ obtain $\vF$ as $\vF=\vG- \til{\vI}^K$. 
Use $\vF$ and the orthonormal 
basis $\vu_i,\,i=1,\ldots,L$ to retrieve 
the embedded numbers ${{h_i,\,i=1,\ldots,L}}$ 
\be
\label{ret}
{h_i} = \la \vu_i , \vF\ra_F,\, i=1,\ldots,L.
\ee
\end{itemize}
\begin{figure}
\begin{center}
\includegraphics[width=4.0cm]{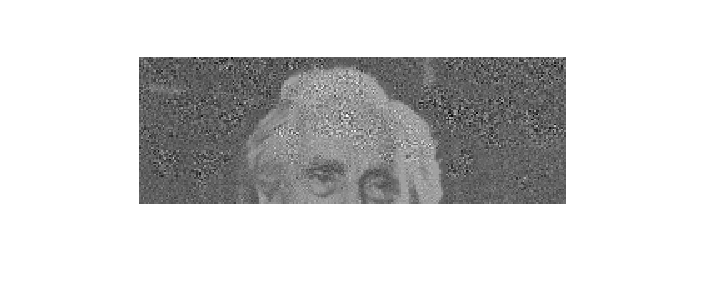}\hspace{-0.5cm}
\includegraphics[width=4.2cm]{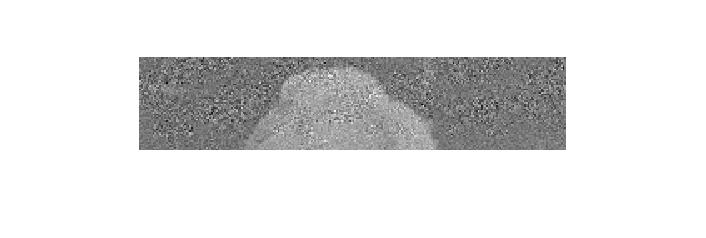}\\
\includegraphics[width=4.2cm]{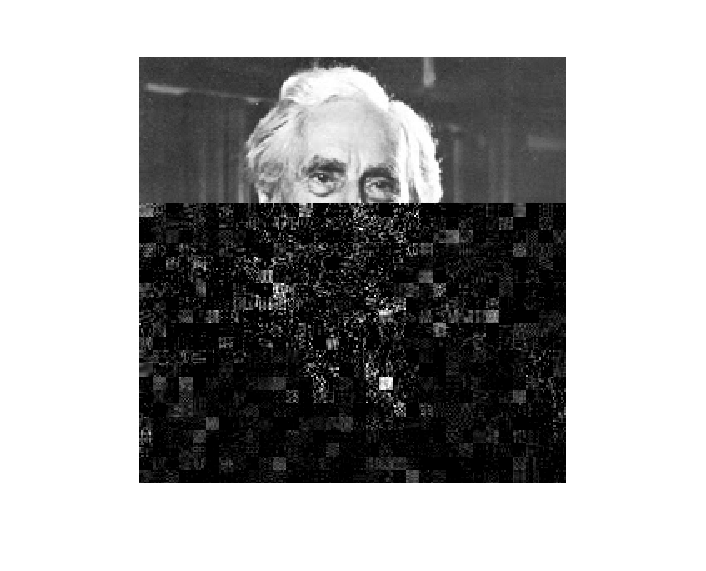}\hspace{-0.5cm}
\includegraphics[width=4.2cm]{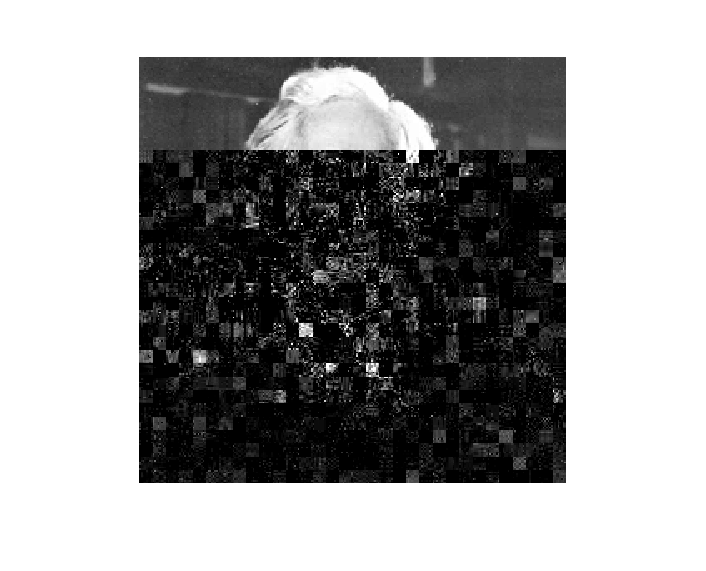}
\includegraphics[width=4.2cm]{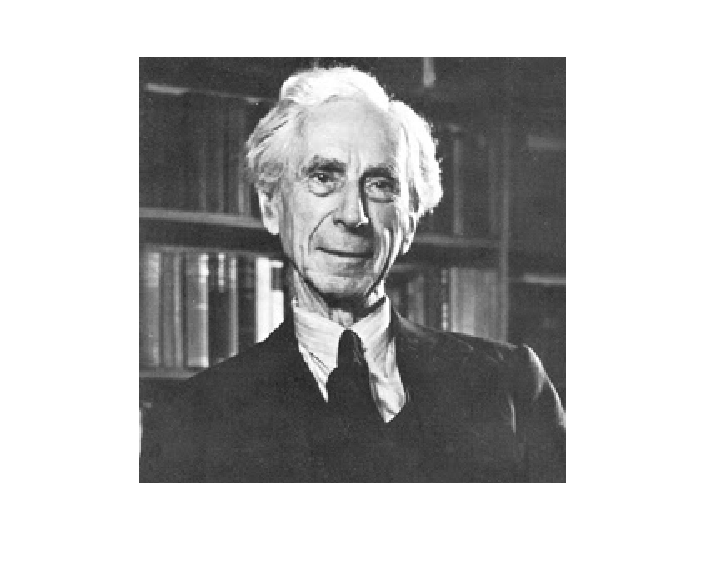}\hspace{-0.5cm}
\includegraphics[width=4.2cm]{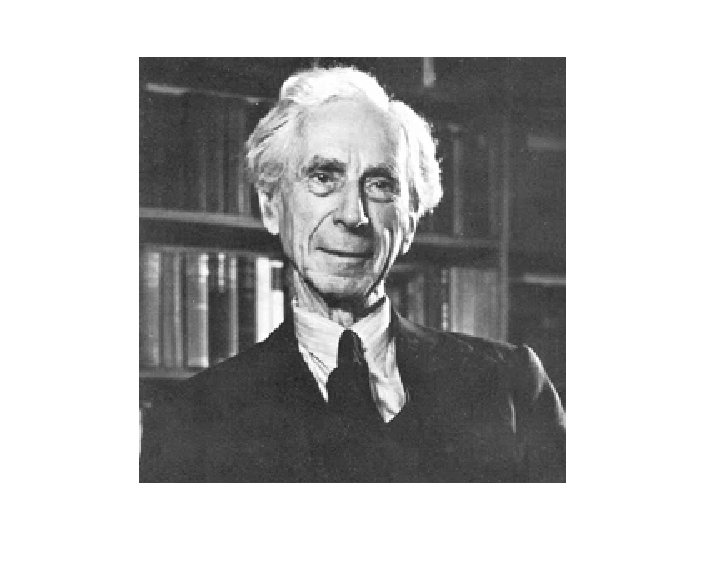}\\
\end{center}
\vspace{-0.85cm}
\caption{The small pictures at the top are the folded Image by 
DCT (left) and RDC-DB dictionary (right).
The middle pictures are the corresponding  
unfolded images without knowledge of 
the private key to initialize the rotation. The bottom 
pictures are the unfolded images when the correct 
${\key}$ is used.}
\label{image_folding}
\end{figure}
One can encrypt the embedding procedure  
simply by randomly controlling the order of the orthogonal 
basis $\vu_i ,\, i=1,\ldots,L$ or by applying some random rotation to 
the basis. An example is 
given in the next section.
\section{Application to Encrypted Image Folding (EIF)}
\label{eif}
We apply now the above discussed embedding scheme to 
fold and encrypt an image. For this we  process the image by 
dividing it into $Q$ blocks $\vI_q,\,q=1,\ldots,Q$
of $N_q\times N_q$ pixels each and compute their 
sparse representation  
\be
\label{repbq}
\vI_q^{K_q}=\sum_{i=1}^{K_q} c_i^{K_q} \vd_{q_{\ell_i}},\,q=1,\ldots,Q.
\ee
We keep a number, $H$, of these block of pixels 
as hosts for embedding 
the coefficients of the remaining equations \eqref{repbq}.  
Each host block $\vI_q^{K_q}$ is embedded as follows:   
Taking $L_q=N_q^2-K_q$ of the coefficients to be embedded, we
build a block of pixels $\vF_q$ as in \eqref{plain} and
add it to the host block to obtain $\vG_q= \vI_q^{K_q} + \vF_q$. 
Since the number $H$ of host blocks is the  superior integer part
of $\frac{Q}{SR}$, as sparsity increases less host 
blocks are needed to embed the remaining ones.
In the example presented here for each host block $q$, 
with $q=1,\ldots,H$, we have built 
the orthogonal basis $\vu_i^q,\,i=1,\ldots,L_q$ (c.f. \eqref{plain})
by randomly generating  matrices
$\vy_i^q \in R^{N_q\times N_q},\,i=1,\ldots,L_q$
using a {{\em public}} initialization ${\seed}_q$ for the random generator.
Through  a  projection matrix ${P}_{\VS_{K_q}}$ onto $\VS_{K_q}=
\Spann\{\vd_{q_{\ell_i}}\}_{i=1}^{K_q}$,
we compute matrices ${\vo}_i^q \in \VS_K^\bot$ as 
\be
\label{vectors}
\vo_i^q= \vy_i^q - P_{\VS_{K_q}} \vy_i^q,\,i=1,\ldots,L_q.  
\ee
Setting an initialization ${\key}$, which remains 
{\em{unknown}} for an unauthorized user, we apply 
a random transformation $\Pi_{{\key}}$ on these matrices
to obtain a private set of matrices
\be
\label{vx}
\Pi_{{\key}}: (\vo_i^q,\,i=1,\ldots,L_q)
\rightarrow \{\vot_i^q\}_{i=1}^{L_q}.
\ee
Next, through an orthogonalization procedure 
$\widehat{\mathbf{Orth}}(\cdot)$ we obtain the orthonormal basis 
\be
\label{vu}
\{\vu_i^q\}_{i=1}^{L_q}= 
\widehat{\mathbf{Orth}}(\vot_i^q,\,i=1,\ldots,L_q).
\ee
that we use for embedding the coefficients of the remaining $Q-H$ blocks.

We illustrate the results on a 8 bit 
$256 \times 256$ photo of Bertrand Russell  
divided into blocks of $8 \times 8$ pixels,
using both standard DCT and the RDC-DB dictionary discussed in 
Sec.~\ref{dictionaries}.

The top pictures of Fig.~\ref{image_folding} are the folded 
images using DCT (left) and the RDC-DB dictionary (right).
Each block of $8 \times 8$ pixels in these figures
is the superposition $\vG_q= \vI_q^{K_q} + \vF_q$ described above.
In both cases the method applied for finding the sparse representation 
$\vI_q^{K_q}$ is nonlinear, but the DCT case is $O(\overline{K})$ faster 
than the mixed dictionaries one ($\overline{K}$ being the 
average number of coefficients per block). 
Since the SR for the DCT is smaller than the SR for the
mixed dictionary, the corresponding folded image is larger.  
The middle pictures are the unfolded images when 
an {\em{incorrect}} security ${\key}$ is used. They are obtained 
as follows: Each block $\vG_q$ in the top pictures is used 
to recover  
the host blocks  $\til{\vI}_q^{K_q},\,q=1,\ldots,H$ as 
$\til{\vI}_q^{K_q}= {P}_{\VS_{K_q}} \vG_q, \,q=1,\ldots,H$
(top piece of image correctly reconstructed).
Subtracting these pixels to the corresponding $\vG_q$ of the 
top picture we obtain the pixels $\vF_q$ which are used to 
retrieve the embedded coefficients, as in \eqref{ret} but with 
 matrices $\vu_i^q,\,i=1,\ldots,L_q$ constructed with an incorrect 
${\key}$
(c.f. \eqref{vu}). As seen in the largest portion of the middle pictures,
with these coefficients the image cannot be reconstructed at all.
The bottom pictures 
are obtained in the same way but using the correct ${\key}$.  
Let us point out that, for reconstructing the image from the coefficients, 
additional space has to be allowed to store the
indices of the atoms in the decomposition \eqref{repbq}.
This is a requirement of nonlinear approximations for general dictionaries. 
\begin{remark}
In order to store the folded image at the same bit depth as the original 
image we need to quantize the blocks of pixels $\vG_q$ to 
convert them into integer numbers, which implies some loss
of information. 
However, the quantization step does not prevent us from recovering 
the coefficients corresponding to the folded pixels with enough accuracy 
to produce a good representation of those blocks of image. 
The PSNR of the recovered image $\til{\vI}^K$ in Fig.~1
(after folding it with the RDC-DB dictionary and subsequent 
rounding) is 
$40.60$ dB while the PSNR of the original approximation $\vI^K$ 
is $40.88$ dB. This implies a relative error due to quantization 
of $0.68\%$. 
In further tests, involving forty five 8 bit images of 
different size and format, the mean value 
relative error due to quantization was $1.32\%$ 
with standard deviation $0.57\%$.
\end{remark}
\begin{remark}
Let us emphasize that since the proposed encryption scheme is based on 
the orthogonal matrices  which define a linear 
transformation (c.f. \eqref{plain}),
as pointed out in \cite{BKW06} it might be vulnerable to 
plain text attack. This means that an attacker could discover the 
matrices $\vu_i^q,\,i=1,\ldots,L_q$ by collecting, for 
each  block, $L_q$-correctly decrypted 
sets of $L_q$ numbers $h_{i,j}^q\,i,j=1,\ldots,L_q$ encrypted with the identical 
 matrices $\vu_i^q,\,i=1,\ldots,L_q$. This would indeed allow to pose $L_q$ equations of 
the form \eqref{plain} and, for 
invertible systems,  disclose the operator used for the encryption. 
However, the initial step for the
construction of this operator involves  matrices
$\vy_i^q,\,i=1,\ldots,L_q$ (c.f. \eqref{vectors}) which are
randomly generated using a public ${\seed}_q$. Hence, 
for a fixed secret $\key$, 
 it is enough to change the public ${\seed}_q$ to avoid  
 $L_q$ encryptions with the identical matrices $\vu_i^q,\,i=1,\ldots,L$.
 Thereby, a simple initial setup for the public random initialization 
 of the encryption process prevents the possibility of plain text attack.
\end{remark}
\section{Conclusions}
\label{conclu}
A bonus of sparse image representation has been discussed: 
the capability for simultaneous storage and encryption 
by simple processing steps.
It was shown that this feature can be used for EIF.
The proposed procedure is applicable through any
appropriate transformation. The example given here has been 
produced by a)DCT and  b)a combination of RDC and DB dictionaries which 
is suitable for image processing by blocking. 
The latter was shown to improve sparsity performance through
nonlinear approximation techniques such as OMP. 
The gain in sparsity also implies that the processing time for the actual
folding and unfolding operations is less in the
RDC-DB case, as it involves less host blocks to be processed. 
On the whole, the time spent in both cases is comparable.
Using a 2.8Ghz AMD processor with 3GB of RAM, the running time
for producing the example of Fig.~1 with MATLAB is 
(average of ten independent runs)
a) 2.39 seconds for DCT and b) 7.05 seconds for RDC-DB. Using a MEX file
implementing OMP in C++ the time of b) is reduced to 1.28 seconds.
These results suggest that advances in matters of
sparse representations may benefit this application.


\subsection*{Acknowledgements}
Support from EPSRC, UK, grant  (EP$/$D062632$/$1) is acknowledged.
We would like to thank Prof Tony Constatinides, from Imperial College,
London, for the enjoyable discussions which inspired and further encouraged
the application of Image Folding. 
The software for reproducing the example
is available from \cite{webpage2}, in section EIFS.
\bibliographystyle{IEEEbib}
\bibliography{revbib}
\end{document}